\pdfoutput=1

\documentclass[11pt]{article}

\usepackage{EMNLP2022}

\usepackage{times}
\usepackage{latexsym}
\usepackage{amssymb}
\usepackage{tabularx}
\usepackage{graphicx}
\usepackage{tabularx}
\usepackage{multirow}
\usepackage{multicol}
\usepackage{caption}
\usepackage{verbatim}
\usepackage{booktabs}
\usepackage{fontawesome5}
\usepackage{hyperref}
\usepackage[noabbrev,capitalize,nameinlink]{cleveref}
\usepackage{tikz}
\usepackage{tikz-dependency}
\usepackage{adjustbox}
\usepackage{todonotes}
\usepackage[multiple]{footmisc}

\newcommand{\bertb}{BERT\textsubscript{base}}
\newcommand{\roberta}{RoBERTa\textsubscript{base}}
\newcommand{\distilbert}{DistilBERT\textsubscript{base}}
\newcommand{\clinbert}{Clinical-BioBERT}
\newcommand{\biobert}{BioBERT}
\newcommand{\twitterbert}{Twitter-RoBERTa\textsubscript{base}}
\newcommand{\scibert}{SciBERT\textsubscript{base}}
\newcommand*\circled[1]{\tikz[baseline=(char.base)]{
            \node[shape=circle,draw,inner sep=.6pt] (char) {#1};}}
\newcommand{\cls}{\texttt{[CLS]}}
\newcommand{\sos}{\texttt{<s>}}
\newcommand{\std}[2]{{#1}{\footnotesize$\pm${#2}}}
\newcommand{\eqcontrib}{\raisebox{.07em}{\resizebox{1em}{!}{\circled{\tiny\faEquals}}}}

\usepackage[T1]{fontenc}

\usepackage[utf8]{inputenc}

\usepackage{microtype}

\usepackage{inconsolata}

\title{Evidence $>$ Intuition: Transferability Estimation for Encoder Selection}

\author{Elisa Bassignana\textsuperscript{\eqcontrib{}}\textsuperscript{\faCompass} \hspace{.5em} Max M\"uller-Eberstein\textsuperscript{\eqcontrib{}}\textsuperscript{\faCompass} \hspace{.5em} Mike Zhang\textsuperscript{\eqcontrib{}}\textsuperscript{\faCompass} \hspace{.5em} Barbara Plank\textsuperscript{\faCompass}\textsuperscript{\faMountain}\textsuperscript{\faRobot}\\
        \textsuperscript{\faCompass}Department of Computer Science, IT University of Copenhagen, Denmark \\ 
        \textsuperscript{\faMountain}Center for Information and Language Processing (CIS), LMU Munich, Germany \\
     \textsuperscript{\faRobot}Munich Center for Machine Learning (MCML), Munich, Germany \\
        \texttt{\{elba, mamy, mikz}\}\texttt{@itu.dk} \hspace{.5em} \texttt{b.plank@lmu.de}}

\begin{document}
\maketitle

\begingroup
\renewcommand\thefootnote{}
\footnotetext{\textsuperscript{\eqcontrib{}} The authors contributed equally to this work.}
\endgroup

\begin{abstract}
\looseness=-1
With the increase in availability of large pre-trained language models (LMs) in Natural Language Processing (NLP), it becomes critical to assess their fit for a specific target task a priori---as fine-tuning the entire space of available LMs is computationally prohibitive and unsustainable. However, \textit{encoder transferability estimation} has received little to no attention in NLP. In this paper, we propose to generate quantitative evidence to predict which LM, out of a pool of models, will perform best on a target task \textit{without} having to fine-tune all candidates. We provide a comprehensive study on LM ranking for 10 NLP tasks spanning the two fundamental problem types of classification and structured prediction. We adopt the state-of-the-art Logarithm of Maximum Evidence (LogME) measure from Computer Vision (CV) and find that 
it positively correlates with final LM performance in 94\% of the setups.
In the first study of its kind, we further compare transferability measures with the de facto standard of human practitioner ranking, finding that evidence from quantitative metrics is more robust than pure intuition and can help identify unexpected LM candidates.
\end{abstract}

\section{Introduction}
Advances in Deep Learning-based NLP and CV build on expressive representations from encoder models pre-trained on massive corpora. Downstream models make use of latent information in these representations to extract relevant features for the task at hand. Within this paradigm, deciding which pre-trained encoder to use in any task-specific architecture is crucial, however training a model using each encoder candidate is infeasible. In absence of prior heuristics (e.g., via related work), the choice of encoder has therefore prevailingly been based on practitioner intuition rather than quantitative evidence.

In NLP, prior work has examined the different yet related task of performance prediction \citep{xia-etal-2020-predicting,ye-etal-2021-towards}, surveyed and categorized LMs~\citep{xia-etal-2020-bert}, and used probing to predict LM performance specifically for dependency parsing~\cite{muller-eberstein-etal-2022-sort},
but has yet to extensively investigate how to rank the increasingly large number of pre-trained LM encoders across various tasks and domains.
Preliminary work by \citet{you2021logme} shows that the LogME estimator holds promise, including the first steps for encoder selection in NLP. With their main focus being on CV, however, they evaluate only a limited set of tasks and models for NLP and use self-reported benchmark scores instead of running controlled experiments which should include, e.g., the variance across initializations, domains, and fine-tuning strategies (\cref{sec:transferability}).
As such, we seek to answer: \textit{How well can we estimate the transferability of pre-trained LMs to specific NLP tasks?}
To do so, we contribute:
\begin{itemize}
    \item The broadest encoder selection study in NLP to date, on 10 domain-diverse classification \textit{and} structured prediction tasks (\cref{sec:experiments});
    \item An extensive evaluation and analysis across multiple dimensions of variation, including seven general vs.\ domain-specific LMs, \cls{} vs.\ mean representations, and head vs.\ full model fine-tuning (\cref{sec:analysis});
    \item A study with NLP experts, comparing the prevailing ranking of LMs by human intuition with LogME's empirical evidence (\cref{sec:human-study}); 
    \item Guidelines for applying and interpreting transferability measures to NLP (\cref{sec:conclusion}), and an open-source toolkit for efficient, task-adaptive LM pre-selection.\footnote{\url{https://github.com/mainlp/logme-nlp}}
\end{itemize}

\begin{table*}[t]
    \centering
    \resizebox{.99\linewidth}{!}{
    \begin{tabular}{l|l|lrrr}
    \toprule
    & \textsc{Dataset}                                 & \textsc{Task} & \textsc{Train / Dev} &  $|\mathcal{Y}|$  & \textsc{Metric} 
    \\
    \midrule
    \multirow{6}{*}{\rotatebox{90}{\textsc{Classification}}}
    & AGNews~\cite{zhang2015character}                 & Topic Classification\          & 84K / 12K                  & 4     & micro-F1      
    \\
    & Airline~\cite{twitter2020kaggle}                 & Sentiment Analysis\            & 10K / 1.5K                 & 3     & micro-F1      
    \\
    & SciERC~\cite{luan-etal-2018-multi}               & Relation Classification        & 1.9K / 275                & 7     & macro-F1      
    \\
    & MNLI~\cite{williams-etal-2018-broad}             & Natural Language Inference     & 393K / 20K                & 3     & micro-F1      
    \\
    & QNLI~\cite{rajpurkar-etal-2016-squad}            & Q\&A/Natural Language Inference & 105K / 5.4K              & 2     & micro-F1      
    \\
    & RTE~\cite{giampiccolo-etal-2007-third}           & Natural Language Inference     & 2.5K / 3K                 & 3     & micro-F1      
    \\
    \midrule   
    \multirow{4}{*}{\rotatebox{90}{\textsc{Str.\ Pred.}}}
    & EWT~\cite{silveira14gold}                     & Dependency Labeling             & 12.5k / 2k            & 36     & micro-F1      
    \\
    & CrossNER~\cite{liu2021crossner}                  & Named Entity Recognition       & 15K / 3.5K 
    & 4     & span-F1      
    \\
    & CrossNER~\cite{liu2021crossner}                  & Named Entity Recognition       & 200 / 450 
    & 17     & span-F1    
    \\
    & JobStack~\cite{jensen-etal-2021-de}              & De-identification              & 18K / 2K  
    & 11    & span-F1       
    \\           
    \bottomrule
    \end{tabular}}
    \caption{\textbf{Datasets.} Indicated are the 10 datasets used in this study, distinguished between the two NLP problem types C and SP for a wide variety of tasks and domains. C tasks cover AGNews (news articles), Twitter Airline Sentiment (Airline; Twitter feedback), SciERC (AI proceedings), MNLI (speech, (non-)fiction, government), QNLI (Wikipedia) and RTE (Wikipedia, news). Within the SP tasks, we experiment on the English Web Treebank (EWT; social media, reviews, emails), CrossNER (news, scientific Wikipedia) and JobStack (Stack Overflow job ads). For each task, we report their \textsc{Train}/\textsc{Dev} split, label space, and task-specific performance metric.}
    \label{tab:datasets}
\end{table*}

\section{Transferability Estimation}\label{sec:transferability}

Transferability estimation aims to quantify the ability of a model to transfer knowledge learned from one task to another ~\cite{eaton2008modeling,sinapov2015learning}. Formally, given a pool of $L$ pre-trained LMs $\{\phi_l\}^L_{l=1}$ and a dataset $\mathcal{D}$, we calculate a predictive score $S_l(\mathcal{D})$ for each  $\phi_l$ which ideally correlates with the model's final performance $P_l(\mathcal{D})$.
$S_l(\mathcal{D})$ is computed without fine-tuning $\phi_l$ on $\mathcal{D}$ such that the optimal $\phi^*_l$ can be chosen from a large model pool at a low computational cost.

The CV community has begun to explore methods for encoder pre-selection and ranking through metrics such as LogME and the Log Expected Empirical Prediction (LEEP; \citealp{nguyen2020leep}). These are widely-used state-of-the-art methods in CV. Recent work introduced the Gaussian Bhattacharyya Coefficient (GBC;~\citealp{pandy2021transferability}) and Optimal Transport based Conditional Entropy (OTCE;~\citealp{tan2021otce}), the exploration of which we leave for future work. However, in the NLP field, related work focus on choosing a task and \emph{not} an LM encoder for transferability~\cite{vu-etal-2020-exploring,padmakumar2022exploring}, leaving the ranking of encoders an unexplored question. 

\paragraph{LogME} LogME measures the suitability of all encoded dataset features $F \in \mathbb{R}^{|D|\times h}$ (e.g., embeddings with dimensionality $h$) to predict all scalar labels $y \in \mathbb{R}^{|D|}$ via the probability density $p(y|F)$. As this density is intractable, it is estimated by mapping $F \to y$ using a linear transformation $w$; this is akin to training a linear probe with optimal parameters $w^*$ and using the likelihood $p(y|F, w^*)$ as a proxy for feature suitability. Because a simple linear model will overfit on the training data, it would be beneficial to obtain the marginal likelihood, or evidence, by integrating over all possible values of $w$: $p(y|F) = \int p(y|F,w)p(w)dw$. To once again make this computation tractable, \citet{you2021logme} reformulate it as an efficient, iterative evidence maximization problem where both $w$ as well as $y$ are drawn from lightly parametrized, isotropic Gaussian distributions. The normalized logarithm of the maximized evidence (LogME) can then be used as $S_l(\mathcal{D})$ to rank encoder models directly.

\paragraph{NLP Setting} 

LogME has shown promise for CV, and an initial study on the GLUE benchmark \citep{wang2018glue} indicate the same for NLP \citep{you2021logme}.
However, for NLP, there are notable differences in setups across tasks. 
We adapt and apply LogME extensively to a wide range of NLP settings to identify empirically grounded guidelines. 

In particular, we investigate variations concerning the task, instance granularity, domain, and tuning strategy. First, compared to most image classification tasks, NLP tasks are subject to differences in granularity, i.e., \textbf{classification} (C) and \textbf{structured prediction} (SP). Furthermore, there is less clarity than for individual images as to which representation best captures the full language input~\cite{mosbach-etal-2020-interplay-fine}. Therefore, for C setups we experiment with two representations: i.e., using \cls{}/\sos{} versus mean over sequence/subwords.

Second, depending on differences in the data domain, NLP practitioners are often faced with a pool of domain-adapted LMs in addition to more general-purpose encoders---the correct choice of which may not be immediately apparent.

Finally, the best performance in NLP is often achieved using full fine-tuning, while CV models usually do not fine-tune the encoder \citep{peters-etal-2019-tune}. It will therefore be crucial to investigate whether the predictive performance of $S_l(\mathcal{D})$ holds when it is computed based on untuned $F$ while $P_l(\mathcal{D})$ is based on fully fine-tuned representations.

\section{Experimental Setup}\label{sec:experiments}

Applying seven architecturally and domain-diverse pre-trained LMs with up to four configurations each to 10 datasets and a wide variety of tasks, we investigate LogME's predictive power for transferability estimation in NLP---for a total of 280 setups. We refer to~\cref{tab:datasets} for our detailed set of tasks.

\paragraph{Language Models} We pick seven pre-trained LMs with a wide domain and architectural variety from the \texttt{Transformers} library's model hub~\cite{wolf2020transformers}. Three are ``general-purpose'' models, namely \bertb{}~\cite{devlin2018bert}, \roberta{}~\cite{liu2019roberta}, and~\distilbert{}~\cite{Sanh2019DistilBERTAD}. Four models are pre-trained on domain-specific corpora, namely \clinbert{}~\cite{alsentzer-etal-2019-publicly}, \biobert{}~\cite{lee2020biobert}, \twitterbert{}~\cite{barbieri-etal-2020-tweeteval}, and \scibert{}~\cite{beltagy-etal-2019-scibert}.
Note that for BioBERT variants domain-adaptive pre-training has been applied~\cite{gururangan2020don}.

\paragraph{Model Setups} The model setup follows the same structure for each task: A pre-trained LM encoder and a 3-layer perceptron head, following~\citet{tenney-etal-2019-bert}. The input to the latter is either the \cls{} token or mean over sequence subwords for C tasks or mean over token subwords for SP tasks. While it is common in CV to keep the encoder frozen and only fine-tune the task-specific head, we also evaluate the practice of full model fine-tuning, as is more common in NLP \citep{peters-etal-2019-tune}. Considering these variations (frozen vs.\ fine-tuning, and \cls{} vs.\ mean), we obtain up to four setups per C task and two setups per SP task. Each experiment is run with five random seeds. Details for reproducibility can be found in~\cref{apx:reproducibility}.

\paragraph{Evaluation} Following~\citet{you2021logme}, we evaluate LogME's predictive power for ranking LMs according to their final performance by using the two correlation coefficients Pearson's $\rho$ and weighted Kendall's $\tau_w$~\citep{vigna2015}, both in $[-1,1]$. Kendall's $\tau_w$ further allows us to estimate the probability of a higher-ranked LM actually performing better by computing $\frac{\tau_w + 1}{2}$.

\begin{figure*}[ht]
    \centering
    \includegraphics[width=.497\linewidth]{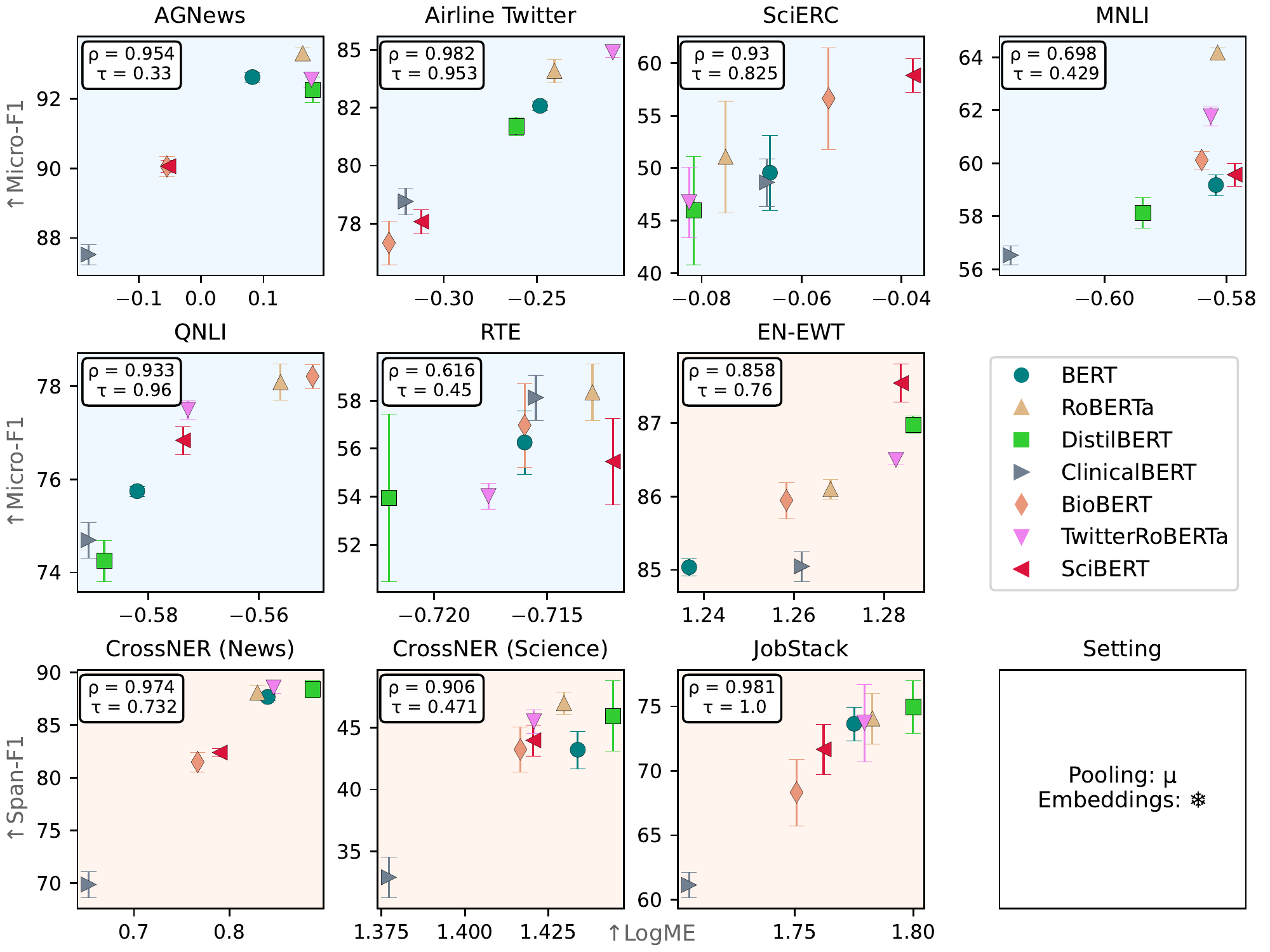}
    \includegraphics[width=.497\linewidth]{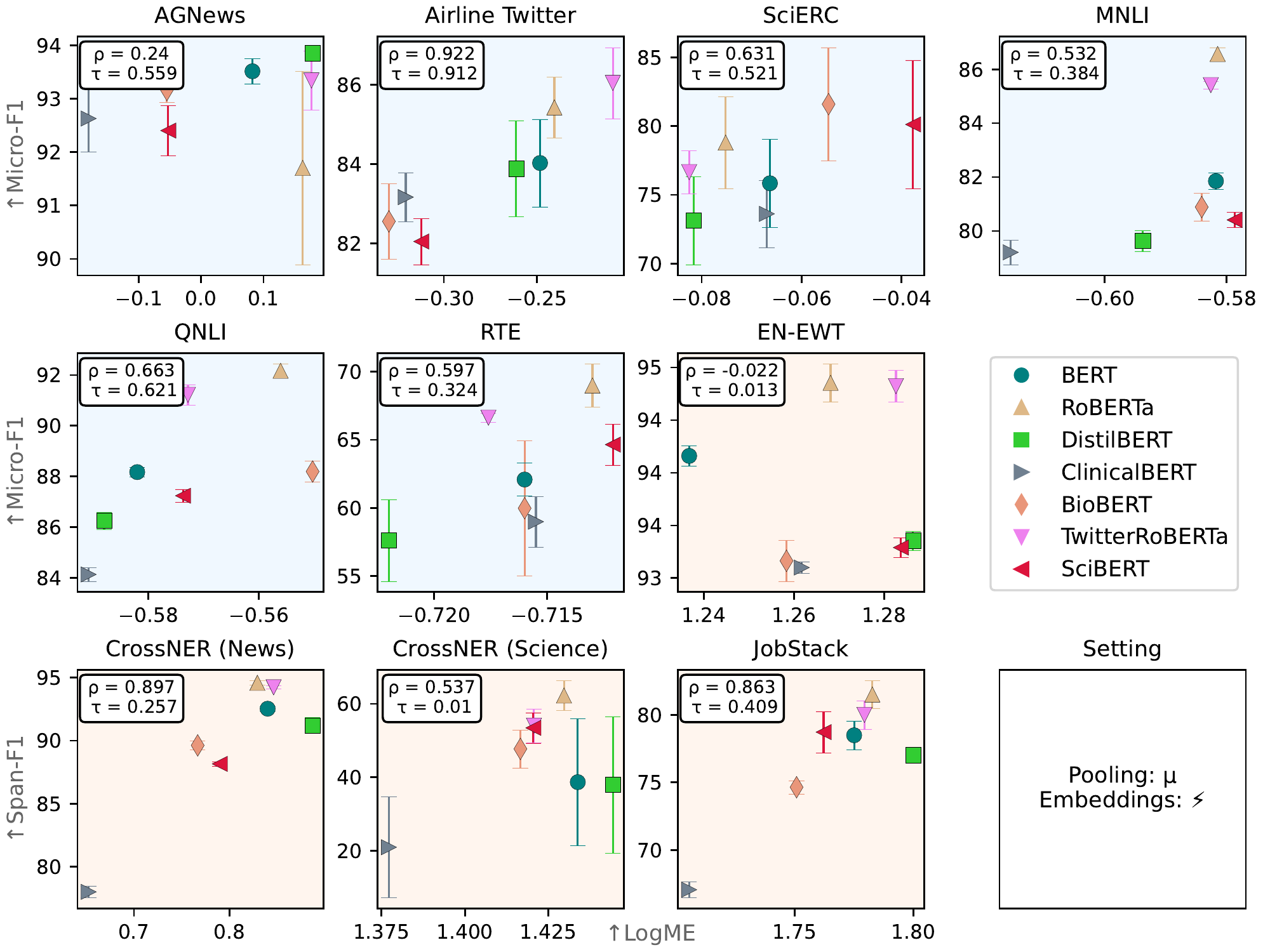}
    \caption{\textbf{Results of Mean Pooling ($\mu$).} We plot the model's LogME scores against their task-specfic performances on each dataset based on mean pooling the token embeddings (\textbf{left}: Frozen embeddings (\faSnowflake), \textbf{right}: Full model fine-tuning (\faBolt)). Task-types are indicated in specific colors: Lightblue for C and beige for SP. Further reported are the Pearson correlation coefficient ($\rho$) and weighted Kendall's tau ($\tau$).}
    \label{fig:results-mean}
\end{figure*}

\begin{figure*}[ht]
    \centering
    \includegraphics[width=.497\linewidth]{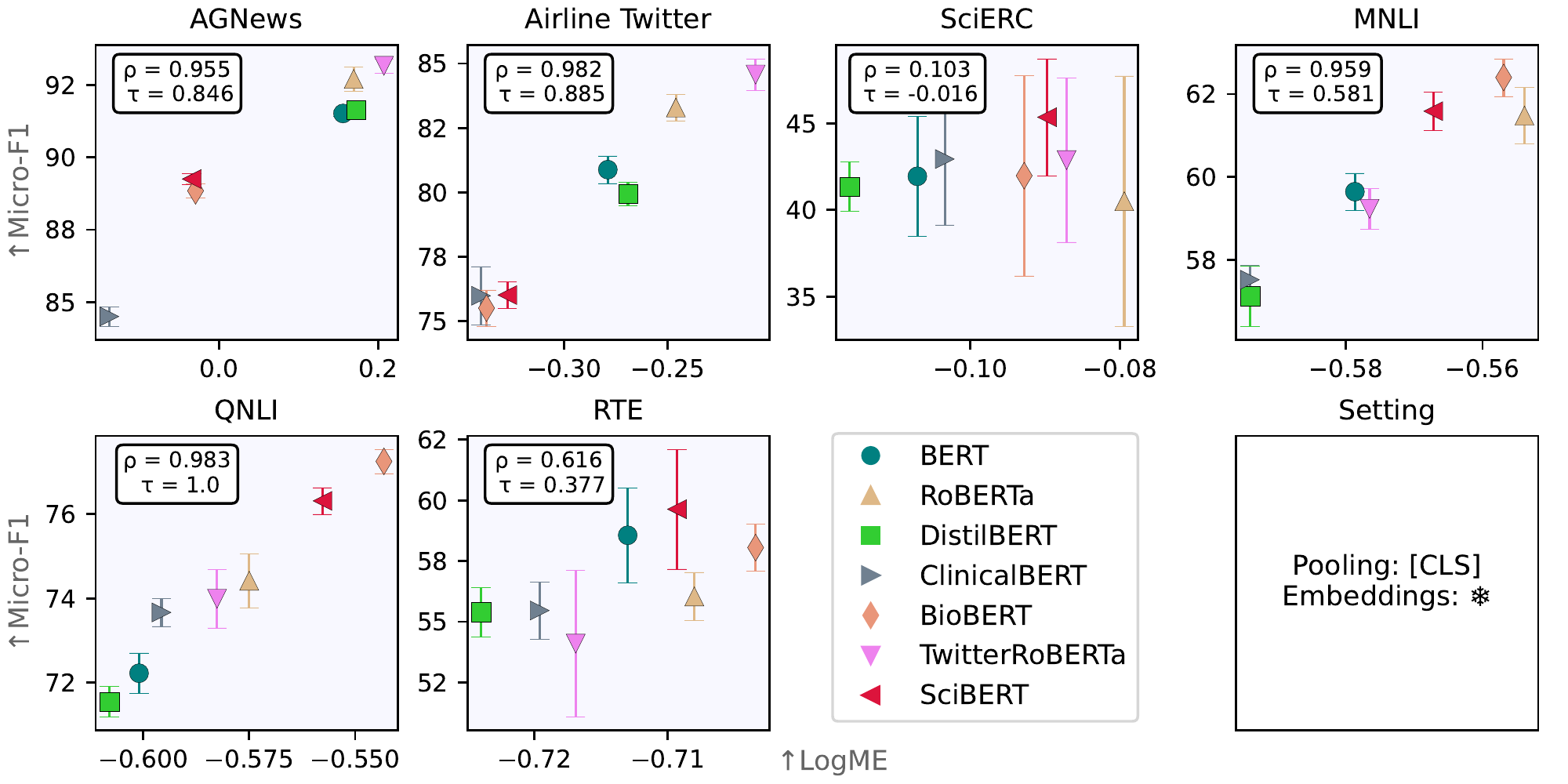}
    \includegraphics[width=.497\linewidth]{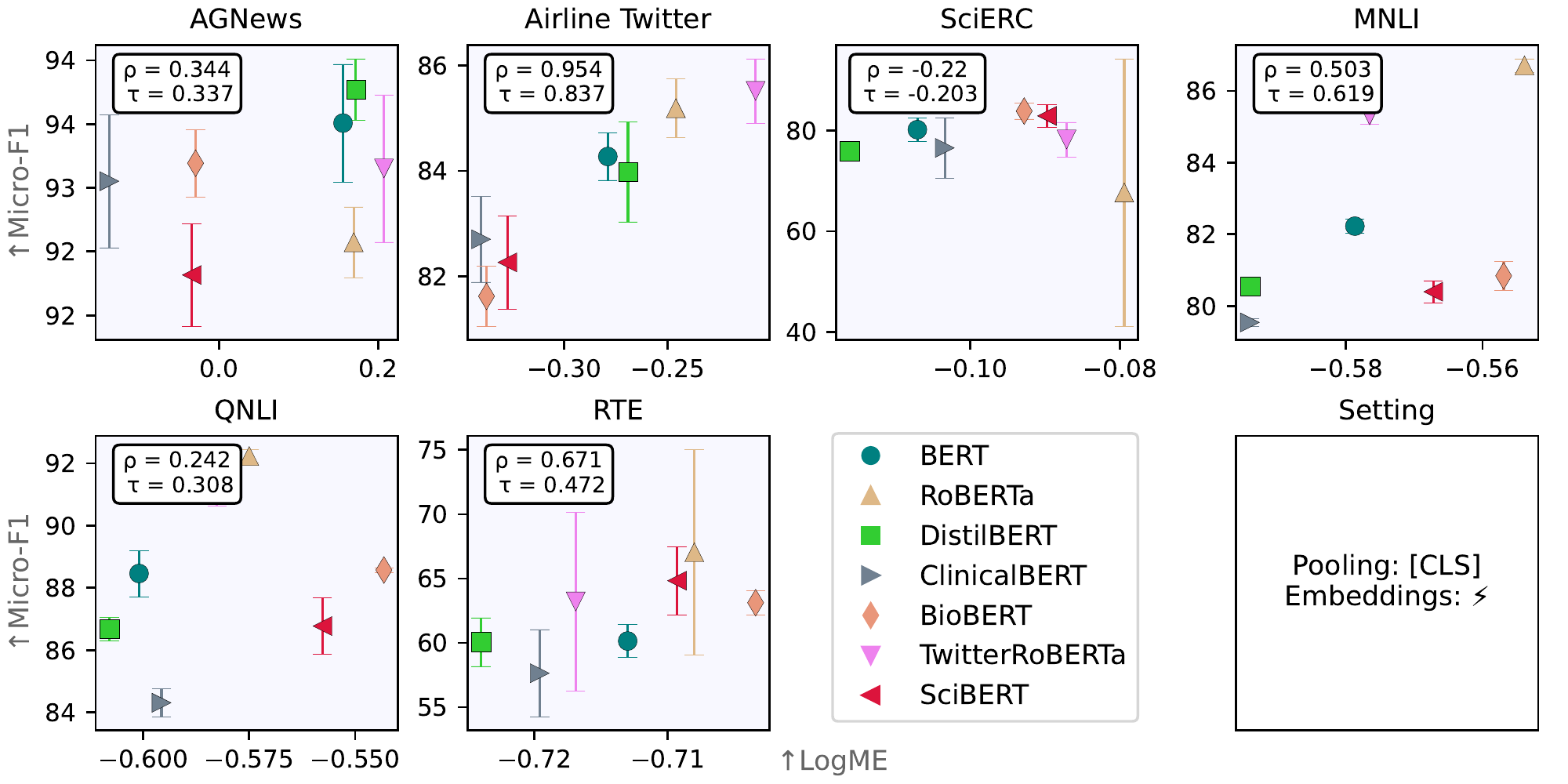}
    \caption{\textbf{Results of \texttt{[CLS]}.} We plot the model's LogME scores against their task-specfic performances on each dataset based on the vector representation of the \texttt{[CLS]} token (\textbf{left}: Frozen embeddings (\faSnowflake), \textbf{right}: Full model fine-tuning (\faBolt)). Reported are the Pearson correlation coefficient ($\rho$) and weighted Kendall's tau ($\tau$).}
    \label{fig:results-cls}
\end{figure*}

\section{Analysis of Results}\label{sec:analysis}
\looseness=-1
Our results across all setups are consolidated in \cref{fig:results-mean} and~\cref{fig:results-cls} (C: blue, SP: beige).\footnote{Exact results can be found in \cref{app:exact}.}\footnote{We successfully reproduce the results reported in \citet{you2021logme} for MNLI, QNLI and RTE.} The left of each figure plots the performance using frozen LM embeddings (\faSnowflake) against LogME scores, while on the right, full LM fine-tuning is applied (\faBolt).\footnote{Note that LogME is only computed on frozen embeddings and does not differ between \faSnowflake{} and \faBolt.}

\looseness=-1
\cref{fig:results-mean} shows the results of using mean-pooled embeddings in both C/SP settings. For \faSnowflake, we obtain $\rho > 0.8$ on 8/10 tasks and $\tau_w > 0.7$ on 6/10 tasks, indicating a strong relationship between model performance and LogME. 
After fine-tuning (\faBolt), we observe a general reduction in $\rho$ and $\tau_w$ (most on CrossNER, EN-EWT), however overall correlations remain positive to a significant degree.

\looseness=-1
For C setups using the alternative \cls{}/\sos{} representations (\cref{fig:results-cls}), LogME correlates highly at $\rho > 0.95$ on 5/6 tasks and $\tau_w > 0.7$ on 4/6 tasks when using head-only tuning (\faSnowflake). After full fine-tuning (\faBolt), SciERC, RTE and AGNews have lower correlations, particularly with the high-variance RoBERTa model. However, the remaining tasks maintain a stable correlation, with $\rho > 0.6$ and $\tau_w > 0.3$ across 5/6 tasks.

\looseness=-1
Overall, LogME has a positive correlation with final performance in 30/32 cases. In more detail, LogME has a $\tau_w > 0.41$ in 20/32 setups, meaning that selecting a higher ranked model is the better choice 71\% of the time. LogME both identifies intuitive, domain-specific scenarios (e.g., Twitter-RoBERTa performing well on Airline Twitter), but also finds cases that may be unintuitive, such as DistilBERT's occasionally high performance for CrossNER and JobStack. This finding holds across C, SP, domains as well as different input representations. For the latter, we note that, surprisingly, even the untuned representation of \cls{}/\sos{} seems to contain useful information with comparable performance to mean pooling.

\looseness=-1
Comparing \faSnowflake{} versus \faBolt, we notice that, as expected, model performance improves, but in general, LogME's predictive power decreases. 
The fully fine-tuned model makes predictions on updated representations such that decreases in predictive performance are inevitable unless the initial LM already represents a local optimum for the task at hand. This fact is crucial for NLP practitioners where full fine-tuning is the standard practice. Taking these factors into account, LogME's efficiency is especially beneficial, as it offers an 86$\times$ speedup over full model fine-tuning \citep{you2021logme}, and its positive correlation in 94\% of our evaluated setups indicates that it is an effective score for transferability estimation in NLP.

\section{Human Performance}\label{sec:human-study}
\looseness=-1
Given the lack of prior work examining transferability estimation of pre-trained LM encoders, the most common method for encoder selection employed today is practitioner intuition. As such, we conduct a small-scale study with 12 NLP practitioners and ask them to perform the same ranking as in \cref{sec:experiments}. Despite having access to model details and associated papers, this task is difficult even for experts. While for LogME, the range of $\tau_w$ is in $[-0.20; 1.00]$, human rankings fall into a wider range of $[-0.54; 1.00]$, indicating higher uncertainty. Similarly, we observe that human correlations are negative thrice as often as for LogME. Additionally, LogME provides a continuous scale for comparing models, while human rankings offer no indication of relative performance differences.  At the same time, they are more inaccurate for tasks without an associated domain-specific model (e.g., news, mixture of genres in EWT). Moreover, even when domains are clear (e.g., Twitter, science), LogME tends to be more accurate than the predictions of most human participants.  Finally, the high variance between practitioners and the fact that no single person was an expert in all setups further reinforces the necessity of quantitative transferability scores.

\section{Conclusion}\label{sec:conclusion}
\looseness=-1
We show the value of transferability estimation for selecting high-performing LMs before full model fine-tuning in experiments, covering the two fundamental NLP settings of classification and structured prediction. By adopting the state-of-the-art LogME scoring method, we are able to rank LMs on a continuous scale which correlates with final performance---with the better encoder being chosen in 71\% of cases. Additionally, we identify NLP-specific guidelines for transferability estimation: In particular, predicting the best LM for tasks/domains which greatly deviate from an encoder's pre-training setup and require large amounts of full fine-tuning may require larger pre-selections of LMs due to the higher uncertainty of the scoring methods. Finally, our human study showed that practitioners frequently misconstrue the performance of LMs even on domain-specific tasks. As such transferability quantification methods provide valuable evidence over intuition.

\section*{Limitations}
A key limitation that practitioners should consider is that, while LogME is viable for the quantitative transferability estimation of LM encoders, there is a noticeable drop in predictive accuracy after full model fine-tuning. We attribute this to the misalignment between the frozen representations of the encoder, which LogME is applied to, and the representations after fine tuning. As stated in \cref{sec:analysis}, unless the untuned LM already constitutes a local optimum for the task at hand, task-specific shifts in its parameters and representations are inevitable.

This similarly applies to cases where the untuned representations differ substantially from what a fully fine-tuned model uses during training. Specifically, for the relation classification task of SciERC, it is important to note that the input given to the model is augmented with special tokens delimiting the entities involved in the relation~\cite{baldini-soares-etal-2019-matching} which are unknown to the untuned model and thus the representations that LogME is computed on. Furthermore, for EN-EWT we suspect that dependency labeling is a more fundamental task solvable with high accuracy by most LMs, especially after fine-tuning as reflected in micro-F1 scores between 93--95. This is mirrored by work on probing untuned LMs which identifies high levels of inherent dependency information \citep{tenney-etal-2019-bert,muller-eberstein-etal-2022-probing}.

Such sensitivity to representational shifts is not exclusive to LogME: In preliminary experiments, we examined LEEP~\cite{nguyen2020leep} as an alternative predictive score $S_l(\mathcal{D})$. Its original use was to rank the transferability of a classifier trained on one dataset, to a new task---leaving the ranking of pre-trained LMs for future work. LEEP has so far only been applied to CV tasks, but we apply it to LM ranking on the collection of NLP tasks above. Our initial experiments achieved low and unintuitive correlations between LEEP's $S_l(\mathcal{D})$ and $P_l(\mathcal{D})$. We speculate that this is due to the absence of a normalizing factor over the number of source classes, i.e., the high number of embedding dimensions in our case (see Equation 2 in \citealp{nguyen2020leep}). While it would further be valuable to investigate methods beyond LEEP and LogME, as mentioned in \cref{sec:transferability}, we leave their evaluation on NLP to future work. At the time of writing, the former two were the most extensively explored in CV, in addition to the original LogME work containing an initial study showing promise for NLP.

Finally, our human ranking study in \cref{sec:human-study} was limited by the number of practitioners with a publication record which we could contact confidentially. However, the group still constituted a diverse set over seniority, gender, and cultural background. A larger group would cover a broader range of backgrounds and may produce different rankings. However, as the surveyed group already displayed high variance, overall predictive performance is unlikely to be significantly higher.

Keeping these limitations in mind, correlations do remain mostly positive for LogME and scores are well suited to be applied to high-dimensional embedding spaces, such that it offers a predictive and efficient measure for quantifying transferability compared to human practitioner intuition.

\section*{Ethics Statement}
\looseness=-1
It is difficult to foresee ethical issues for this work due to the broad applicability of LM encoder pre-selection. To the best of our knowledge, in the CV community from which our evaluated scoring methods originate, there have been no harmful applications thus far. In fact, as fine-tuning the entire space of available language models is unsustainable and unethical in terms of climate sustainability, efficient encoder pre-selection methods such as LogME provide a positive first step towards tackling this problem.

\section*{Acknowledgements}
\looseness=-1
We would like to wholeheartedly thank the members NLPnorth group at the IT University of Copenhagen, The Center for Information and Language Processing at the Ludwig Maximilian University of Munich, and the Dialogue Modeling Group at the University of Amsterdam for comments on earlier versions of this work and participating in the human practitioner ranking study.

MZ and BP are supported by the Independent Research Fund Denmark (DFF) grant 9131-00019B. EB, MME, and BP are supported by the Independent Research Fund Denmark (DFF) Sapere Aude grant 9063-00077B. BP is supported by the ERC Consolidator Grant DIALECT  101043235.

\bibliography{anthology,custom}
\bibliographystyle{acl_natbib}

\appendix

\section*{Appendix}
\label{sec:appendix}

\section{Reproducibility}
\label{apx:reproducibility}

Each model is trained on an NVIDIA A100 GPU with 40GBs of VRAM and an AMD Epyc 7662 CPU. The seed numbers the models are initialized with are 4012, 5060, 9908, 8857, 8823. We run the models for 30 epochs with a patience of 3 on each respective dev.\ data. We use a batch size of 16, 32, or 64 depending on the size of the dataset. When keeping the language model weights frozen, we use a learning rate of 1$e$--3. For full model fine-tuning, the learning rate is set at 5$e$--5. On GLUE, JobStack, and CrossNER (News), we observed training instability and set the learning rate to 5$e$--7. The evaluated LMs have between 66M parameters (\distilbert) and 125M parameters (\roberta), taking between 10 minutes (SciERC) and 3 days (e.g., AGNews, MNLI) to fully fine-tune. Keeping the LM frozen and only fine-tuning the task-specific head is around 70\% more time-efficient. Computing LogME requires one forward pass to embed the dataset instances, before completing the score calculation in under 1 minute.

\begin{table*}
\centering
\resizebox{\textwidth}{!}{
\begin{tabular}{c|r|rrr|rrr}
\toprule
& & \multicolumn{3}{c|}{$\mu$} & \multicolumn{3}{c}{\cls{}} \\
\midrule
 \textsc{Dataset} & \textsc{Language Model} & \textsc{LogME} & \textsc{Frozen} & \textsc{Tuned} & \textsc{LogME} & \textsc{Frozen} & \textsc{Tuned} \\
\midrule
\multirow{7}{*}{\rotatebox{90}{AGNews}} 
& \texttt{bert-base-uncased}                  & 0.0822  & \std{92.62}{0.13} & \std{93.51}{0.23} & 0.1555  & \std{91.52}{0.10} &\std{93.51}{0.46} \\
& \texttt{roberta-base}                       & 0.1628  & \std{93.30}{0.17} & \std{91.70}{1.81} & 0.1689  & \std{92.71}{0.42} &\std{92.57}{0.28} \\
& \texttt{distilbert-base-uncased}            & 0.1786  & \std{92.26}{0.37} & \std{93.85}{0.11} & 0.1716  & \std{91.65}{0.22} &\std{93.77}{0.24} \\
& \texttt{emilyalsentzer/Bio\_ClinicalBERT}   & -0.1801 & \std{87.52}{0.29} & \std{92.62}{0.62} & -0.1384 & \std{84.50}{0.34} &\std{93.05}{0.52} \\
& \texttt{dmis-lab/biobert-v1.1}              & -0.0548 & \std{90.05}{0.29} & \std{93.16}{0.23} & -0.0300 & \std{88.84}{0.25} &\std{93.19}{0.26} \\
& \texttt{cardiffnlp/twitter-roberta-base}    & 0.1768  & \std{92.55}{0.21} & \std{93.34}{0.55} & 0.2070  & \std{93.16}{0.25} &\std{93.15}{0.58} \\
& \texttt{allenai/scibert\_scivocab\_uncased} & -0.0527 & \std{90.06}{0.17} & \std{92.40}{0.47} & -0.0348 & \std{89.25}{0.20} &\std{92.32}{0.40} \\\hline
& $\rho$, $\tau_w$                              &         & 0.954, 0.330      & 0.240, 0.559      &         &  0.955, 0.846     & 0.344, 0.337   \\
\midrule
\multirow{7}{*}{\rotatebox{90}{Airline}} 
& \texttt{bert-base-uncased}                  & -0.2484 & \std{82.58}{0.19} & \std{84.03}{1.10} & -0.2789 & \std{80.88}{0.53} & \std{84.27}{0.45} \\
& \texttt{roberta-base}                       & -0.2407 & \std{84.10}{0.52} & \std{85.43}{0.77} & -0.2460 & \std{83.29}{0.51} & \std{85.19}{0.56} \\
& \texttt{distilbert-base-uncased}            & -0.2612 & \std{81.71}{0.39} & \std{83.89}{1.21} & -0.2691 & \std{79.95}{0.45} & \std{83.99}{0.95} \\
& \texttt{emilyalsentzer/Bio\_ClinicalBERT}   & -0.3205 & \std{78.46}{0.57} & \std{83.17}{0.62} & -0.3402 & \std{75.98}{1.14} & \std{82.70}{0.81} \\
& \texttt{dmis-lab/biobert-v1.1}              & -0.3295 & \std{76.67}{0.93} & \std{82.55}{0.96} & -0.3376 & \std{75.50}{0.71} & \std{81.62}{0.57} \\
& \texttt{cardiffnlp/twitter-roberta-base}    & -0.2094 & \std{84.89}{0.21} & \std{86.05}{0.90} & -0.2074 & \std{84.57}{0.62} & \std{85.51}{0.61} \\
& \texttt{allenai/scibert\_scivocab\_uncased} & -0.3122 & \std{77.58}{0.51} & \std{82.05}{0.58} & -0.3275 & \std{76.01}{0.51} & \std{82.27}{0.88} \\\hline
& $\rho$, $\tau_w$                              &         & 0.982, 0.953     & 0.922, 0.912     &         & 0.982, 0.885      & 0.954, 0.837     \\
\midrule
\multirow{7}{*}{\rotatebox{90}{SciERC}} 
& \texttt{bert-base-uncased}                  & -0.0663 & \std{49.56}{3.55} & \std{75.84}{3.21} & -0.1071 & \std{41.94}{3.45} & \std{80.20}{2.37} \\
& \texttt{roberta-base}                       & -0.0752 & \std{51.07}{5.34} & \std{78.80}{3.34} & -0.0794 & \std{40.51}{7.22} & \std{67.71}{26.6} \\
& \texttt{distilbert-base-uncased}            & -0.0816 & \std{45.98}{5.17} & \std{73.13}{3.20} & -0.1161 & \std{41.35}{1.43} & \std{75.95}{1.93} \\
& \texttt{emilyalsentzer/Bio\_ClinicalBERT}   & -0.0669 & \std{48.64}{2.26} & \std{73.61}{2.44} & -0.1034 & \std{42.94}{3.80} & \std{76.57}{6.06} \\
& \texttt{dmis-lab/biobert-v1.1}              & -0.0546 & \std{56.64}{4.85} & \std{81.60}{4.13} & -0.0928 & \std{41.98}{5.77} & \std{83.89}{1.58} \\
& \texttt{cardiffnlp/twitter-roberta-base}    & -0.0825 & \std{46.75}{3.35} & \std{76.65}{1.58} & -0.0871 & \std{42.87}{4.73} & \std{78.25}{3.46} \\
& \texttt{allenai/scibert\_scivocab\_uncased} & -0.0377 & \std{58.83}{1.61} & \std{80.12}{4.68} & -0.0897 & \std{45.35}{3.38} & \std{82.93}{2.30} \\\hline
& $\rho$, $\tau_w$                            &         & 0.930, 0.825      & 0.631, 0.521      &         & 0.103, -0.016    & -0.220, -0.203 \\
\midrule
\multirow{7}{*}{\rotatebox{90}{MNLI}} 
& \texttt{bert-base-uncased}                  & -0.5818 & \std{59.18}{0.39} & \std{81.85}{0.31} & -0.5786 & \std{59.64}{0.45} & \std{82.23}{0.19} \\
& \texttt{roberta-base}                       & -0.5815 & \std{64.18}{0.19} & \std{86.57}{0.24} & -0.5539 & \std{61.48}{0.68} & \std{86.71}{0.19} \\
& \texttt{distilbert-base-uncased}            & -0.5938 & \std{58.13}{0.58} & \std{79.64}{0.39} & -0.5940 & \std{57.13}{0.73} & \std{80.54}{0.09} \\
& \texttt{emilyalsentzer/Bio\_ClinicalBERT}   & -0.6154 & \std{56.53}{0.35} & \std{79.21}{2.44} & -0.5940 & \std{57.52}{0.45} & \std{79.54}{0.11} \\
& \texttt{dmis-lab/biobert-v1.1}              & -0.5841 & \std{60.12}{0.33} & \std{80.89}{4.13} & -0.5569 & \std{62.40}{0.51} & \std{80.84}{0.41} \\
& \texttt{cardiffnlp/twitter-roberta-base}    & -0.5826 & \std{61.77}{0.36} & \std{85.41}{1.58} & -0.5765 & \std{59.23}{0.13} & \std{85.32}{0.25} \\
& \texttt{allenai/scibert\_scivocab\_uncased} & -0.5787 & \std{59.57}{0.44} & \std{80.41}{4.68} & -0.5672 & \std{61.59}{0.28} & \std{80.40}{0.31} \\\hline
& $\rho$, $\tau_w$                            &         & 0.698, 0.429      & 0.532, 0.384      &         & 0.959, 0.581      & 0.503, 0.619\\
\midrule
\multirow{7}{*}{\rotatebox{90}{QNLI}} 
& \texttt{bert-base-uncased}                  & -0.5823 & \std{75.75}{0.11} & \std{88.17}{0.19} & -0.6008 & \std{72.23}{0.48} & \std{88.46}{0.74} \\
& \texttt{roberta-base}                       & -0.5557 & \std{78.09}{0.39} & \std{92.17}{0.26} & -0.5749 & \std{74.42}{0.64} & \std{92.23}{0.22} \\
& \texttt{distilbert-base-uncased}            & -0.5881 & \std{74.25}{0.44} & \std{86.26}{0.33} & -0.6079 & \std{71.55}{0.36} & \std{86.68}{0.38} \\
& \texttt{emilyalsentzer/Bio\_ClinicalBERT}   & -0.5908 & \std{74.69}{0.38} & \std{84.13}{0.27} & -0.5957 & \std{73.67}{0.33} & \std{84.31}{0.46} \\
& \texttt{dmis-lab/biobert-v1.1}              & -0.5502 & \std{78.21}{0.26} & \std{88.19}{0.42} & -0.5432 & \std{77.25}{0.29} & \std{88.57}{0.07} \\
& \texttt{cardiffnlp/twitter-roberta-base}    & -0.5728 & \std{77.49}{0.20} & \std{91.22}{0.41} & -0.5826 & \std{73.99}{0.69} & \std{91.03}{0.39} \\
& \texttt{allenai/scibert\_scivocab\_uncased} & -0.5737 & \std{76.84}{0.30} & \std{87.24}{0.26} & -0.5577 & \std{76.31}{0.32} & \std{86.77}{0.90} \\\hline
& $\rho$, $\tau_w$                            &         & 0.933, 0.960      & 0.663, 0.621      &         & 0.983, 1.000      & 0.242, 0.308 \\
\midrule
\multirow{7}{*}{\rotatebox{90}{RTE}} 
& \texttt{bert-base-uncased}                  & -0.7160 & \std{56.26}{1.28} & \std{62.09}{1.21} & -0.7131 & \std{58.56}{1.96} & \std{60.14}{1.28} \\
& \texttt{roberta-base}                       & -0.7133 & \std{58.35}{8.00} & \std{68.99}{1.58} & -0.7081 & \std{56.04}{1.00} & \std{67.05}{8.00} \\
& \texttt{distilbert-base-uncased}            & -0.7220 & \std{53.96}{1.89} & \std{57.63}{3.00} & -0.7237 & \std{55.40}{1.02} & \std{60.07}{1.89} \\
& \texttt{emilyalsentzer/Bio\_ClinicalBERT}   & -0.7155 & \std{58.13}{0.94} & \std{58.99}{1.86} & -0.7196 & \std{55.46}{1.18} & \std{57.64}{3.41} \\
& \texttt{dmis-lab/biobert-v1.1}              & -0.7160 & \std{56.97}{1.74} & \std{59.98}{4.95} & -0.7034 & \std{58.05}{0.97} & \std{63.12}{0.95} \\
& \texttt{cardiffnlp/twitter-roberta-base}    & -0.7176 & \std{54.02}{0.53} & \std{66.63}{0.34} & -0.7169 & \std{54.10}{3.03} & \std{63.21}{6.95} \\
& \texttt{allenai/scibert\_scivocab\_uncased} & -0.7121 & \std{55.46}{1.79} & \std{64.65}{1.51} & -0.7093 & \std{59.64}{2.47} & \std{64.83}{2.66} \\\hline
& $\rho$, $\tau_w$                            &         & 0.616, 0.450      & 0.597, 0.324      &         & 0.616, 0.377      & 0.671, 0.472    \\
\bottomrule
\end{tabular}}
\caption{\textbf{Exact Results of Classification Tasks.} We indicate the \textsc{LogME} score of each model (\textsc{Language Model}) and its performance on a wide variety of datasets (\textsc{Dataset}) in different settings (\textsc{Frozen}, \textsc{Tuned}) by either taking the representations of the tokens and apply mean pooling ($\mu$) or the representation of the \texttt{[CLS]} token. Given the LogME scores and the performance metrics, we can calculate the Pearson correlation coefficient ($\rho$) and the weighted Kendall's tau ($\tau_w$).}
\label{tab:classification}
\end{table*}

\begin{table*}
\centering
\resizebox{.8\textwidth}{!}{
\begin{tabular}{c|r|rrr}
\toprule
& & \multicolumn{3}{c}{$\mu$} \\
\midrule
 \textsc{Dataset} & \textsc{Language Model} & \textsc{LogME} & \textsc{Frozen} & \textsc{Tuned} \\
\midrule
\multirow{7}{*}{\rotatebox{90}{EN-EWT}} 
& \texttt{bert-base-uncased}                  & 1.2367 & \std{85.04}{0.12} & \std{94.16}{0.10} \\
& \texttt{roberta-base}                       & 1.2681 & \std{86.10}{0.13} & \std{94.85}{0.18} \\
& \texttt{distilbert-base-uncased}            & 1.2864 & \std{86.98}{0.12} & \std{93.36}{0.09} \\
& \texttt{emilyalsentzer/Bio\_ClinicalBERT}   & 1.2617 & \std{85.05}{0.20} & \std{93.10}{0.06} \\
& \texttt{dmis-lab/biobert-v1.1}              & 1.2583 & \std{85.95}{0.25} & \std{93.16}{0.19} \\
& \texttt{cardiffnlp/twitter-roberta-base}    & 1.2826 & \std{86.50}{0.07} & \std{94.82}{0.15} \\
& \texttt{allenai/scibert\_scivocab\_uncased} & 1.2837 & \std{87.54}{0.26} & \std{93.29}{0.09} \\\hline
& $\rho$, $\tau_w$                            &        & 0.858, 0.760      & -0.022, 0.013\\
\midrule
\multirow{7}{*}{\rotatebox{90}{CrossNER (News)}} 
& \texttt{bert-base-uncased}                  & 0.8397 & \std{87.66}{0.33} & \std{92.53}{0.17} \\
& \texttt{roberta-base}                       & 0.8290 & \std{88.08}{0.65} & \std{94.59}{0.17} \\
& \texttt{distilbert-base-uncased}            & 0.8867 & \std{88.41}{0.79} & \std{91.21}{0.64} \\
& \texttt{emilyalsentzer/Bio\_ClinicalBERT}   & 0.6527 & \std{69.86}{1.26} & \std{78.01}{0.47} \\
& \texttt{dmis-lab/biobert-v1.1}              & 0.7666 & \std{81.48}{0.92} & \std{89.63}{0.35} \\
& \texttt{cardiffnlp/twitter-roberta-base}    & 0.8460 & \std{88.55}{0.53} & \std{94.23}{0.13} \\
& \texttt{allenai/scibert\_scivocab\_uncased} & 0.7897 & \std{82.38}{0.39} & \std{88.16}{0.18} \\\hline
& $\rho$, $\tau_w$                            &        & 0.974, 0.732      & 0.897, 0.257\\
\midrule
\multirow{7}{*}{\rotatebox{90}{CrossNER (Sci.)}} 
& \texttt{bert-base-uncased}                  & 1.4339 & \std{43.22}{1.51} & \std{38.68}{17.3} \\
& \texttt{roberta-base}                       & 1.4297 & \std{47.00}{0.90} & \std{62.27}{4.02} \\
& \texttt{distilbert-base-uncased}            & 1.4444 & \std{45.96}{2.85} & \std{37.97}{18.6} \\
& \texttt{emilyalsentzer/Bio\_ClinicalBERT}   & 1.3772 & \std{32.89}{1.66} & \std{20.96}{13.8} \\
& \texttt{dmis-lab/biobert-v1.1}              & 1.4166 & \std{43.24}{1.81} & \std{47.73}{5.17} \\
& \texttt{cardiffnlp/twitter-roberta-base}    & 1.4207 & \std{45.51}{0.94} & \std{54.05}{4.61} \\
& \texttt{allenai/scibert\_scivocab\_uncased} & 1.4205 & \std{43.98}{1.24} & \std{53.44}{4.13} \\\hline
& $\rho$, $\tau_w$                            &        & 0.906, 0.471      &  0.537, 0.010 \\
\midrule
\multirow{7}{*}{\rotatebox{90}{JobStack}} 
& \texttt{bert-base-uncased}                  & 1.7750 & \std{73.64}{1.30} & \std{78.49}{1.06}\\
& \texttt{roberta-base}                       & 1.7827 & \std{74.06}{1.96} & \std{81.51}{1.02}\\
& \texttt{distilbert-base-uncased}            & 1.7998 & \std{74.96}{2.03} & \std{77.02}{0.34}\\
& \texttt{emilyalsentzer/Bio\_ClinicalBERT}   & 1.7056 & \std{61.13}{0.99} & \std{67.07}{0.60}\\
& \texttt{dmis-lab/biobert-v1.1}              & 1.7508 & \std{68.32}{2.58} & \std{74.65}{0.49}\\
& \texttt{cardiffnlp/twitter-roberta-base}    & 1.7793 & \std{73.72}{3.00} & \std{79.99}{1.06}\\
& \texttt{allenai/scibert\_scivocab\_uncased} & 1.7621 & \std{71.66}{1.93} & \std{78.72}{1.54}\\\hline
& $\rho$, $\tau_w$                            &        & 0.981, 1.000      & 0.863, 0.409 \\
\bottomrule
\end{tabular}}
\caption{\textbf{Exact Results of Structured Prediction Tasks.} We indicate the \textsc{LogME} score of each model (\textsc{Language Model}) and its performance on a wide variety of datasets (\textsc{Dataset}) in different settings (\textsc{Frozen}, \textsc{Tuned}) by taking the representations of the tokens and apply mean pooling ($\mu$). Here we do not take the representation of the \texttt{[CLS]} token as this has no meaning for the structured prediction task. Given the LogME scores and the performance metrics, we can calculate the Pearson correlation coefficient ($\rho$) and the weighted Kendall's tau ($\tau_w$).}
\label{tab:structpred}
\end{table*}

\section{Exact Results}\label{app:exact}
In \cref{tab:classification} and \cref{tab:structpred}, we present the exact performance numbers shown in \cref{fig:results-mean} and \cref{fig:results-cls}. The results here are separated by task.

\end{document}